\documentclass[sigconf]{acmart}

\copyrightyear{2025}
\acmYear{2025}
\setcopyright{acmlicensed}
\acmConference[CIKM '25] {Proceedings of the 34th ACM International Conference on Information and Knowledge Management}{ November 10--14, 2025}{Seoul, Republic of Korea.}
\acmBooktitle{Proceedings of the 34th ACM International Conference on Information and Knowledge Management (CIKM '25), November 10--14, 2025, Seoul, Republic of Korea}
\acmISBN{979-8-4007-2040-6/2025/11}
\acmDOI{10.1145/3746252.3761541.}

\usepackage{multirow}
\usepackage{graphicx}
\usepackage{svg}
\def\method{\textsc{OASIS}}
\begin{document}

%\title{SSIS: Utilizing Generative Adversarial Network For Sea Salinity Imputation Only By Drifter Trajectories}
\title{OASIS: Harnessing Diffusion Adversarial Network for Ocean Salinity Imputation using Sparse Drifter Trajectories}

\author{Bo Li}
\authornote{Both authors contributed equally to this work.}
\orcid{0009-0007-4822-9858}
\affiliation{%
  \institution{Griffith University}
  \city{Gold Coast}
  % \state{QLD}
  \country{Australia}
}
\email{bo.li6@griffithuni.edu.au}

\author{Yingqi Feng}
\authornotemark[1]
\orcid{0000-0002-0269-8006}
\affiliation{%
  \institution{Florida Atlantic University}
  \city{Boca Raton}
  % \state{FL}
  \country{USA}
}
\email{yfeng2016@fau.edu}

% \author{Ming Jin}
% \orcid{0000-0002-6833-4811}
% \affiliation{%
%   \institution{School of Information and Communication Technology, Griffith University}
%   \city{Brisbane}
%   \state{QLD}
%   \country{Australia}
% }
% \email{ming.jin@griffith.edu.au}

\author{Ming Jin}
\orcid{0000-0002-6833-4811}
\author{Xin Zheng}
\orcid{0000-0003-0915-7787}
\affiliation{%
  \institution{Griffith University}
  \city{Gold Coast}
  \country{Australia}
  % \state{QLD}
  \country{}
}
\email{{ming.jin, xin.zheng}@griffith.edu.au}
% \email{xin.zheng@griffith.edu.au}

% \author{Yufei Tang}
% \orcid{0000-0002-6915-4468}
% \affiliation{%
%   \institution{Florida Atlantic University}
%   % \city{Boca Raton}
%   \state{FL}
% \email{tangy@fau.edu}

\author{Yufei Tang}
\orcid{0000-0002-6915-4468}
\author{Laurent Cherubin}
\orcid{0000-0001-9475-2593}
\affiliation{%
  \institution{Florida Atlantic University}
  \city{Boca Raton, Fort Pierce}
  % \state{FL}
  \country{USA}
}
\email{{tangy, lcherubin}@fau.edu}
% \email{lcherubin@fau.edu}

% \author{Alan Wee-Chung Liew}
% \orcid{0000-0001-6718-7584}
% \affiliation{%
%   \institution{School of Information and Communication Technology, Griffith University}
%   \city{Gold Coast}
%   \state{QLD}
%   \country{Australia}
% }
% \email{a.liew@griffith.edu.au}

\author{Alan Wee-Chung Liew}
\orcid{0000-0001-6718-7584}
\author{Can Wang}
\orcid{0000-0002-2890-0057}
\affiliation{%
  \institution{Griffith University}
  \city{Gold Coast}
  % \state{QLD}
  \country{Australia}
}
\email{{a.liew, can.wang}@griffith.edu.au}
% \email{can.wang@griffith.edu.au}

\author{Qinghua Lu}
\affiliation{%
  \institution{Data61, CSIRO}
  \orcid{0000-0002-9466-1672}
  \city{Alexandria}
  % \state{NSW}
  \country{Australia}
}
\email{qinghua.lu@data61.csiro.au}

% \author{Jingwei Yao}
% \affiliation{%
%   \institution{School of Engineering and Built Environment, Griffith University}
%   \city{Gold Coast}
%   \state{QLD}
%   \country{Australia}
% }
% \email{jingwei.yao@griffithuni.edu.au}

\author{Jingwei Yao}
\affiliation{%
  \institution{Griffith University}
  \orcid{0009-0005-2170-7544}
  \city{Gold Coast}
  % \state{QLD}
  \country{Australia}
}
\email{jingwei.yao@griffithuni.edu.au}
% \email{hong.zhang@griffith.edu.au}

\author{Shirui Pan}
\orcid{0000-0003-0794-527X}
\author{Hong Zhang}
\orcid{0000-0002-2642-5467}
\affiliation{%
  \institution{Griffith University}
  \city{Gold Coast}
  % \state{QLD}
  \country{Australia}
}
\email{{s.pan,hong.zhang}@griffith.edu.au}

\author{Xingquan Zhu}
\orcid{0000-0003-4129-9611}
\affiliation{%
  \institution{Florida Atlantic University}
  \city{Boca Raton}
  % \state{FL}
  \country{USA}
}
\email{xzhu3@fau.edu}

\renewcommand{\shortauthors}{Bo Li et al.}

\begin{abstract}
Ocean salinity plays a vital role in circulation, climate, and marine ecosystems, yet its measurement is often sparse, irregular, and noisy, especially in drifter-based datasets. Traditional approaches, such as remote sensing and optimal interpolation, rely on linearity and stationarity, and are limited by cloud cover, sensor drift, and low satellite revisit rates. While machine learning models offer flexibility, they often fail under severe sparsity and lack principled ways to incorporate physical covariates without specialized sensors. In this paper, we introduce the \textbf{O}ce\textbf{A}n \textbf{S}alinity \textbf{I}mputation \textbf{S}ystem (\textbf{\method}), a novel diffusion adversarial framework designed to address these challenges by: (1) employing a transformer‐based global dependency capturing module to learn long-range spatiotemporal correlations from sparse trajectories; (2) constructing a generative imputation model that conditions on easily observed tidal covariates to progressively refine imputed salinity fields; and (3) using a scheduler diffusion method to enhance the model's robustness. This unified architecture exploits the periodic nature of tidal signals as a proxy for unmeasured physical drivers, without the need for additional equipment. We evaluate~\method~on four benchmark datasets, including one real-world measurement from Fort Pierce Inlet, Florida, USA and three simulated Gulf of Mexico trajectories. Results show consistent improvements over both traditional and neural baselines, achieving up to 52.5\% reduction in MAE compared to Kriging. We also develop a lightweight, web-based deployment system that enables salinity imputation through interactive and batch interfaces, available at: https://github.com/yfeng77/OASIS.
\end{abstract}

\begin{CCSXML}
<ccs2012>
   <concept>
       <concept_id>10010147.10010257.10010293</concept_id>
       <concept_desc>Computing methodologies~Machine learning approaches</concept_desc>
       <concept_significance>500</concept_significance>
       </concept>
   <concept>
       <concept_id>10010147.10010341.10010342.10010343</concept_id>
       <concept_desc>Computing methodologies~Modeling methodologies</concept_desc>
       <concept_significance>500</concept_significance>
       </concept>
   <concept>
       <concept_id>10002951.10003227.10003236</concept_id>
       <concept_desc>Information systems~Spatial-temporal systems</concept_desc>
       <concept_significance>300</concept_significance>
       </concept>
   <concept>
       <concept_id>10010405.10010432.10010437</concept_id>
       <concept_desc>Applied computing~Earth and atmospheric sciences</concept_desc>
       <concept_significance>100</concept_significance>
       </concept>
 </ccs2012>
\end{CCSXML}

\ccsdesc[500]{Computing methodologies~Machine learning approaches}
\ccsdesc[500]{Computing methodologies~Modeling methodologies}
\ccsdesc[300]{Information systems~Spatial-temporal systems}
\ccsdesc[100]{Applied computing~Earth and atmospheric sciences}

\keywords{Missing Data, Time Series, Imputation, Diffusion Adversarial Network, Salinity, Drifter}
\maketitle

\section{Introduction}
Salinity, along with temperature and pressure, is a fundamental parameter for understanding the physical and chemical processes of the ocean~\cite{veronis1972properties}. It plays a key role in driving oceanic phenomena such as thermohaline circulation, water mass formation, and nutrient transport~\cite{schmidt2004links,bray1988water,whitney2005physical}. Therefore, understanding the spatial and temporal variability of salinity is essential for advancing marine science and managing marine resources.

Salinity in the open ocean is relatively conservative, primarily influenced by evaporation, precipitation, and currents~\cite{yu2011global, zheng2024online, zheng2024gnnevaluator}. In contrast, offshore and coastal waters exhibit significant salinity variability driven by freshwater inputs from river discharge, which dynamically interacts with tides, wind forcing, and human activities such as dam operations and land-use changes~\cite{rothig2023human}.  Due to their ecological sensitivity, economic importance, and high population density, these regions are particularly affected by salinity dynamics, which have direct implications for estuarine ecosystems, marine biodiversity, water quality, and the resilience of coastal infrastructure~\cite{smyth2016effects}. Therefore, this study focuses on the \textbf{salinity dynamics of nearshore waters} to improve our understanding of their controlling mechanisms and variability. It addresses the urgent need for robust and practical salinity estimation tools tailored to such complex nearshore environments. 

Drifters equipped with hydrological sensors can be used to measure salinity along the drifter track. However, measurements obtained are often extremely sparse due to the separation between drifter trajectories. As a result, nearshore salinity modeling is naturally framed as an imputation task: estimating missing values at unobserved times and locations from limited, irregular drifter data. Previous studies have developed sea salinity imputation frameworks leveraging remote sensing and optimal interpolation techniques~\cite{cutolo2024cloinet, jin2024spatial, melnichenko2016optimum, melnichenko2014spatial}. However, these approaches assume spatial stationarity and linear dynamics, failing to capture fine-scale temporal variability due to cloud cover, sensor drift, and the limited revisit frequency of satellite instruments. Additionally, satellite-derived salinity is often noisy and depends heavily on in situ calibration and validation~\cite{hormann2015evaluation, centurioni2015sea}. Consequently, it is crucial to construct accurate imputation models capable of robust salinity interpolation. Recent data-driven approaches, from multilayer perceptrons (MLPs)~\cite{liu2013artificial} to deep neural models~\cite{song2020novel, jia2022prediction}, have been applied to salinity forecasting. Some incorporate covariates such as tides~\cite{guillou2023predicting}, yet most focus on forecasting rather than tackling sparse imputation or explicitly modeling spatiotemporal dependencies.

As promising generative models for imputation, Generative Adversarial Network (GAN) models have advanced substantially by learning to map random noise vectors to plausible data distributions. In the imputation context, GANs model the conditional distribution of missing values given observed data, allowing the generator to infer realistic and coherent imputations, while the discriminator distinguishes between observed and imputed entries. This architecture has inspired numerous imputation methods that leverage adversarial training to reconstruct missing values~\cite{yoon2020gamin, wang2021pc, awan2021imputation, deng2022extended, guo2024experimentalevaluationimputationmodels}, making it an effective learning model for addressing the significant data sparsity challenge. Simultaneously, transformer architectures characterized by multi‐head self‐attention demonstrate exceptional capability in capturing global spatiotemporal dependencies, making them well suited for tasks requiring long‐range interactions~\cite{voita2019analyzing}. However, existing GAN-based imputation models often struggle with two key limitations in ocean applications: (1) insufficient capacity to capture long-range dependencies across sparse drifter tracks, and (2) instability or bias due to heterogeneous data distributions caused by drifting sensors and dynamic ocean conditions. These limitations motivate our design of a more robust, physically grounded imputation model tailored to coastal ocean salinity reconstruction.

In this work, we propose the \textbf{O}ce\textbf{A}n \textbf{S}alinity \textbf{I}mputation \textbf{S}ystem (\textbf{\method}), a diffusion adversarial framework that integrates: \textbf{(1) Normalization Module} to remove distributional shifts across drifting sensors, ensuring stable training and reducing bias; \textbf{(2) Global Dependency Capturing (GDC):} a transformer‐based module that learns long‐range spatial–temporal correlations from sparse drifter tracks; \textbf{(3) Scheduler Diffusion Adversarial Network (DAN):} a generative adversarial network refined by a cosine‐schedule noise diffuser and adversarial loss, conditioned on tidal height to progressively enhance imputation fidelity. Our key contributions are: \\
    (1) To the best of our knowledge, this work presents the first integration of generative sea salinity imputation into a fully automated framework tailored for sparse drifter trajectories. \\
    (2) Propose a unified architecture that addresses domain-specific challenges, such as sensor drift, data heterogeneity, and long-range dependencies, by integrating normalization, attention-based modeling, adversarial refinement, and a lightweight deployment component.\\
    (3) Compile four datasets spanning both simulated and real-world drifter trajectories with salinity data, covering diverse conditions and sparsity levels. These serve as benchmarks for ocean monitoring and data-driven environmental research.\\
    (4) Extensive experiments show that \method~consistently outperforms existing baselines. On the real-world dataset, it reduces RMSE by 21.3\% and MAPE by 18.5\% over the best-performing baseline, demonstrating its robustness in realistic nearshore settings.\\
    (5) Develop a lightweight, extensible web-based system that enables real-time salinity imputation through interactive input and batch processing, enhancing accessibility for marine researchers and practitioners.

%The remainder of this paper is organized as follows. Section~\ref{sec:related} reviews related work. Section~\ref{sec:method} details the~\method~framework. Section~\ref{sec:experiments} presents experimental results and analysis.

% The ``\verb|acmart|'' document class can be used to prepare articles
% for any ACM publication --- conference or journal, and for any stage
% of publication, from review to final ``camera-ready'' copy, to the
% author's own version, with {\itshape very} few changes to the source.

\section{Related Work}\label{sec:related}
\subsection{Traditional Statistical Interpolation}
Early ocean salinity imputation relied on Optimal Interpolation (OI) and Kriging methods, which effectively smoothed out the noise by maximizing the likelihood of fusing the observations with the background field. For OI, Fu et al. \cite{fu2011application} describe the implementation of an Ensemble Optimal Interpolation (EnOI) in a two-way nested North/Baltic Sea model for assimilating temperature and salinity profiles. In terms of Kriging, Chen et al. \cite{chen2016optimization} used a Kriging-based interpolation method to estimate the error of the Trophic State Index in Quanzhou Bay.
\subsection{GAN-based Imputation}
Missing data issues pervade a wide range of application domains, highlighting the extensive utility of data imputation models. Furthermore, several studies have leveraged industrial datasets to rigorously evaluate the performance of GAN-based imputation methods \cite{yoon2020gamin, hwang2019hexagan, wang2021pc, awan2021imputation, deng2022extended}. Traffic data imputation represents one of the most widely adopted applications of data imputation techniques. Guo et al. \cite{guo2024experimentalevaluationimputationmodels} evaluated several GAN-based models on spatiotemporal traffic datasets. Moreover, other studies have extended GAN-based imputation to domains such as finance \cite{le2021a2gan}, network measurement \cite{tan2021packet}, cyber-attack analysis~\cite{chawla2023deep}, and air quality monitoring \cite{zhou2021federated}. 
\subsection{Time Series Imputation}
Time series imputation has long been a prominent research area, exploiting both temporal dependencies and inter-feature relationships to restore missing observations. Early approaches relied on statistical methods, such as last-observation carried forward and k-nearest neighbors imputation, to fill data gaps \cite{troyanskaya2001missing, beretta2016nearest}. Furthermore, a substantial body of work has leveraged Recurrent Neural Networks (RNNs) \cite{kim2018temporal, yoon2018deep, jin2023expressive, liu2024self} and Transformer-based architectures \cite{zhao2023transformed} to model temporal dependencies across diverse application domains~\cite{zheng2025test, zheng2024structure, shen2025understanding}, including healthcare \cite{zhang2023improving}, urban analytics \cite{li2022fine}, environmental monitoring \cite{jiang2020bilstm}, and indoor systems \cite{li2023data}.
\begin{figure*}[t]
    \centering
    \includegraphics[width=0.7\textwidth]{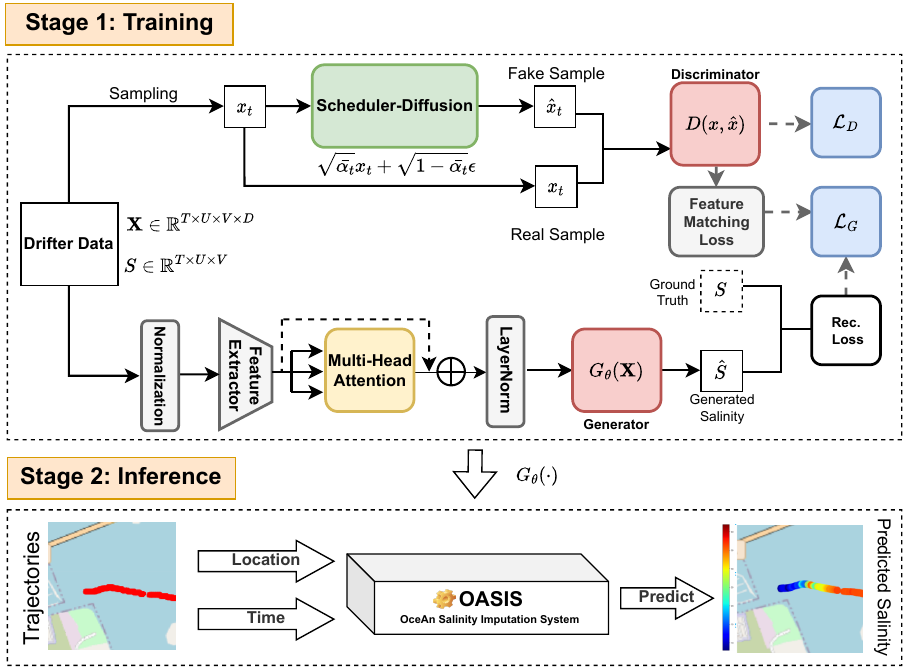}
    \caption{An overview of the two-stage pipeline for~\method~is detailed in Section~\ref{sec:method}: Stage 1 involves the model training process of~\method. For the \textsc{generator}, it involves a normalization module described in Section~\ref{sec: norm}, a Global Dependency Capturing module described in Section~\ref{sec: GDC}, and \textsc{generator} loss formulated by Eq.~\ref{gloss}. In terms of \textsc{discriminator}, it involves \textsc{scheduler diffusion} module described in Section~\ref{sec:GAN} and \textsc{discriminator} loss formulated by Eq.~\ref{dloss}. Stage 2 is inference. For any time and location,~\method~can give an imputed sea salinity value.}
    \label{fig:framework}
\end{figure*}

% In addition to specifying the {\itshape template style} to be used in
% formatting your work, there are a number of {\itshape template parameters}
% which modify some part of the applied template style. A complete list
% of these parameters can be found in the {\itshape \LaTeX\ User's Guide.}

% Frequently-used parameters, or combinations of parameters, include:
% \begin{itemize}
% \item {\texttt{anonymous,review}}: Suitable for a ``double-anonymous''
%   conference submission. Anonymizes the work and includes line
%   numbers. Use with the \texttt{\string\acmSubmissionID} command to print the
%   submission's unique ID on each page of the work.
% \item{\texttt{authorversion}}: Produces a version of the work suitable
%   for posting by the author.
% \item{\texttt{screen}}: Produces colored hyperlinks.
% \end{itemize}

% This document uses the following string as the first command in the
% source file:
% \begin{verbatim}
% \documentclass[sigconf,authordraft]{acmart}
% \end{verbatim}

\section{Methodology}\label{sec:method}
The objective of sea salinity imputation is to reconstruct missing salinity values from sparse and irregularly sampled drifter trajectories. This task is crucial for supporting subsequent oceanographic analysis, particularly in coastal and offshore regions where data is limited and noisy. Traditional approaches, such as Kriging-based interpolation~\cite{chen2016optimization} and remote sensing retrievals, are constrained by strong stationary assumptions, low temporal resolution, and limited accuracy in coastal zones~\cite{kim2023remote}. Statistical methods typically capture only spatial correlations and fail to model spatiotemporal interactions. Meanwhile, satellite-derived salinity products suffer from cloud occlusion, sensor drift, and infrequent revisits, and they are poorly suited to reconstructing fine-grained, transient ocean features. Moreover, drifter observations result in highly sparse and irregular data tensors, posing additional challenges to conventional interpolation schemes. Our ~\method~ overcomes these challenges in two ways. First, we integrate a transformer-based module into our architecture to capture global dependencies within the data (Section~\ref{sec: GDC}). Second, we employ a DAN model to enrich feature representations, reconstruct fine-grained local details, and perform the imputation task (Section~\ref{sec:GAN}).

\subsection{Problem Statement}
\textbf{Drifter Data Representation.} We represent our drifter data as a partially observed 4D tensor \(\mathbf{X} \in \mathbb{R}^{T\times U\times V\times D}\), where $T$ is the number of discrete time steps \(\{t_1,t_2,\cdots,t_T\}\), $U$, $V$ are grid dimensions in latitude and longitude, and $D$ is the number of sensor variables (e.g., salinity, temperature, pressure). The entry \(\mathbf{X}(t,u,v,d)\) is the observed point of one drifter at time \(t\) and spatial cell \((u,v)\) or \(\mathrm{NaN}\) if missing. A corresponding mask \(\mathbf{M}\in\{0,1\}^{T\times V\times U\times D}\) indicates whether each entry is present (1) or missing (0) due to buoy sparsity.

\noindent\textbf{Spatiotemporal Sea Salinity Imputation.} We consider \(K\) drifter trajectories that record \(D\) different sensor variables (e.g.\ tide height, salinity) over \(T\) discrete time-steps \(\{t_1,\dots,t_T\}\).  
To enable joint spatiotemporal imputation, we overlay the study region with a regular grid of size \(U\times V\) (latitude \(\times\) longitude) and aggregate all observations onto this grid.  
The result is a partially observed fourth-order tensor \(\mathbf{X}\;\in\;\mathbb{R}^{T\times U\times V\times D}\). Our goal is to \emph{impute salinity values at arbitrary times and locations} throughout the study area using a model trained on drifter-observed data, denoted as:
\begin{equation}
    \hat{S} = f_\theta 
    \left(t,u,v,d
    \right),
\end{equation}
where \(\hat{S}\in\mathbb{R}^{T\times U\times V}\) is the imputed salinity value at arbitrary times and locations in the study region. %and $t$, $u$, $v$, and $d$ are the time, latitude, longitude, and potential covariate features.

\subsection{Data Normalization}\label{sec: norm}
Although our study concentrates on a specific region, marine environmental conditions exhibit significant heterogeneity across different latitudes and longitudes. To mitigate this distribution shift, we employ the RevIN \cite{kim2022reversible} to statistically influence in drifter data. For each drifter trajectory with \(M\) observations, we compute the mean \(\mu\) and variance \(\sigma^2\) as
\begin{equation}
    \mu = \frac{1}{M} \sum_{i=1}^{M} \mathbf{X}_i,\quad
    \sigma^2 = \frac{1}{M} \sum_{i=1}^{M} (\mathbf{X}_i - \mu)^2.
\end{equation}
We then normalize each observation \(\mathbf{X}\) and apply a learnable affine transformation:
\begin{equation}\label{RevIN}
    \mathbf{X} = \gamma \left(\frac{\mathbf{X} - \mu}{\sqrt{\sigma^2} + \epsilon}\right) + \beta,
\end{equation}
where \(\gamma\) and \(\beta\) are initialized to 1 and 0, respectively, and are optimized during model training.

\subsection{\method: Sea Salinity Imputation Under Sparse Drifter Data}
%The objective of the \method~model is to learn comprehensive representations that simultaneously capture temporal and spatial dependencies while integrating relevant covariates across the study region. The constituent modules of \method~are detailed below.

\subsubsection{Global Dependencies Capturing}\label{sec: GDC}
For sea salinity imputation, the available inputs include temporal information, spatial locations (\(\text{latitude}\times \text{longitude}\)) and some observed covariate features we can get easily, like tide and depth. To cope with the first challenge which cannot be solved by general sea salinity imputation, we propose to utilize transformer-based module to capture the global temporal dependencies and in-channel interactions in the input feature sequences, which mainly consists of Positional Encoding (PE) and Multi-Head Self-Attention (MHA), and combines residual concatenation with layer normalization to stabilize the training. Since self-attention is inherently insensitive to sequence order~\cite{hahn2020theoretical}, positional encoding needs to be injected for each time step. To incorporate positional information into each drifter embedding \(\mathbf{X}\), we first augment it with a positional encoding \(\mathrm{PE}\in\mathbb{R}^{N\times D}\), where \(N=U\times V\):
\begin{equation}
\mathbf{X} = \mathbf{X} + \mathrm{PE}.
\end{equation}
We then project \(\mathbf{X}\) into query, key, and value vectors using three learnable weight matrices \(W_Q\), \(W_K\), and \(W_V\):
\begin{equation}
Q = W_Q \mathbf{X}, 
\quad 
K = W_K \mathbf{X}, 
\quad 
V = W_V \mathbf{X}.
\end{equation}
Finally, we compute scaled dot-product attention as
\begin{equation}
\mathrm{Attention}(Q,K,V) \;=\; \mathrm{softmax}\left(\tfrac{QK^\top}{\sqrt{d_k}}\right)V.
\end{equation}
Hence, the attention score between position \(i\) and \(j\) is:
\begin{equation}
    \text{A}_{i,j}=\frac{\text{exp}(Q_iK_j/\sqrt{d_k})}{\sum_{j'=1}^N\text{exp}(Q_iK_{j'}/\sqrt{d_k})},\quad \mathbf{X}_i=\sum_{j=1}^{N}\text{A}_{i,j}V_i,
\end{equation}
which is a global dependency for position \(i\) to all position \(j=\{1,2...,N\}\).

\subsubsection{DAN-based Sea Salinity Imputation Model}\label{sec:GAN}
Another challenge in sea salinity imputation based on drifter data is the inherent sparsity of observations. Because drifter trajectories do not provide dense coverage of the study area, we employ a DAN that refines GAN to mitigate this issue. On the one hand, GANs have demonstrated effectiveness in addressing data sparsity issues~\cite{alzubaidi2023survey}. On the other hand, since the underlying distribution within a specific study region is expected to be relatively homogeneous, GANs can be trained to learn an ideal generative model capable of accurately imputing missing values~\cite{goodfellow2014generative}.

In this section, the salinity imputation model based on the diffusion adversarial framework is proposed. The whole model consists of three core components: (1) a diffusion noise scheduler (\textsc{Scheduler Diffusion}), (2) a discriminator (\textsc{Discriminator}), and (3) a generator (\textsc{Generator}). During the training process, the \textsc{Diffusion} module is responsible for injecting multi-scale noise into real and fake samples, while the \textsc{Discriminator} and \textsc{Generator} are alternately optimized under the driving force of adversarial and feature matching, which ultimately enables the \textsc{Generator} to accurately impute salinity values.

\textbf{Scheduler Diffusion.} We use a cosine scheduling strategy to design the forward diffusion process. Let \(T\) be the total diffusion steps, so at step t, the weight at this step would be \(\beta_t=\beta_0+\frac{1}{2}(\beta_T-\beta_0)(1+\text{cos}(\pi\frac{T-t}{T}))\), \(\alpha_t=1-\beta_t\), \(\bar{\alpha}_t=\prod_{i=1}^t\alpha_i\). Hence, for any input \(x_t\), through noise \(\epsilon\sim\mathcal{N}(0,I)\), we can get the noisy sample as:
\begin{equation}\label{schedular}
    \hat{x_t}=\sqrt{\bar{\alpha_t}}x_t+\sqrt{1-\bar{\alpha_t}}\epsilon
\end{equation}
This process provides multi-scale noise perturbations for subsequent adversarial training, enhancing the robustness of the model.

\textbf{Discriminator.} \textsc{Discriminator} \(D\) is encouraged to distinguish real and fake samples after \textsc{Scheduler Diffusion}, the \textsc{discriminator} loss is defined as:
\begin{equation}\label{dloss}
    \mathcal{L}_D=\frac{1}{2}(\text{BCE}(D(\hat{x}),0)+\text{BCE}(D(x),1))
\end{equation}
where \(\text{BCE}(\cdot)\) is binary cross entropy.

\textbf{Generator.} \textsc{Generator} $G_\theta$ is designed to learn complex correlations among spatiotemporal inputs, covariates, and salinity, enabling accurate imputation. After reversible normalization and positional encoding, the input $\mathbf{X}$ is processed by a multi-level linear feature extractor (FE) and an attention module (Attn), yielding
\begin{equation}
\hat{s} = G_\theta
\left(\mathrm{Attn}\bigl(\mathrm{FE}\bigl(\gamma\tfrac{\mathbf{X}-\mu}{\sqrt{\sigma^2}+\epsilon} + \beta\bigr)\bigr)\right).
\end{equation}
The \textsc{generator}’s objective comprises two components: \textbf{Reconstruction loss}, which drives the generated salinity $\hat{s}_i$ toward the true value $s_i$. \textbf{Feature-matching loss}, which encourages the \textsc{discriminator}’s hidden activations for fake samples to align with those of real samples. Formally,
\begin{equation}\label{gloss}
\mathcal{L}_G 
= \underbrace{\frac{1}{N}\sum_{i=1}^N (s_i - \hat{s}_i)^2}_{\text{MSE term}}
+ \underbrace{\frac{1}{M}\sum_{j=1}^M \lVert f_{\mathrm{real},j} - f_{\mathrm{fake},j}\rVert_1}_{\text{feature-matching term}},
\end{equation}
where $f_{\mathrm{real}}$ and $f_{\mathrm{fake}}$ denote the \textsc{discriminator}’s feature vectors at a chosen hidden layer for real and generated samples, respectively.

\textbf{Adversarial Objective.} The overall training objective is formulated as a minimax game between the \textsc{generator} $G$ and the \textsc{discriminator} $D$:
\begin{equation}
\min_{G}\,\max_{D}\;V(D, G),
\end{equation}
where
\begin{equation}
V(D, G)
=
\underbrace{\mathbb{E}_{x\sim p_{\mathrm{data}}}\bigl[\log D(x)\bigr]}_{\substack{\text{Encourage }D\text{ to correctly}\\\text{classify real samples}}}
\;+\;
\underbrace{\mathbb{E}_{X\sim p_X,\,t}\bigl[\log\bigl(1 - D\bigl(q_t\bigl(G(X)\bigr)\bigr)\bigr)\bigr]}_{\substack{\text{Encourage }D\text{ to correctly}\\\text{identify generated samples}}}\,.
\end{equation}

\begin{table*}[hbtp]
\centering
\caption{Datasets characteristics}
\vspace{-0.3cm}
\label{tab:dataset}
\begin{tabular}{lccccc}
\toprule
\textbf{Datasets}   & \textbf{\# of  trajectories} & \textbf{\# of time steps} & \textbf{Datetime} & \textbf{Lon / Lat Range (°)} & \textbf{Salinity Range (psu)} \\ \midrule
\textbf{FP Observed} & 4 & 11,426 & 16 Jun. 2016 & [-80.32, -80.26] / [27.46, 27.47] & [06.82, 42.00] \\
\textbf{GoM-10}   & 53 & 21,847 & Oct. 2021 & [-91.63, -82.81] / [26.01, 30.34] & [30.12, 36.51] \\
\textbf{GoM-11}   & 41 & 9878  & Nov. 2021 & [-91.63, -82.65] / [26.01, 30.34] & [29.29, 36.13]\\ 
\textbf{GoM-12}  & 28 & 13,573 & Dec. 2021 & [-91.63, -82.65] / [26.01, 30.21] & [32.90, 36.30]\\ 
\bottomrule   
\end{tabular}
\vspace{-0.3cm}
\end{table*}

% Modifying the template --- including but not limited to: adjusting
% margins, typeface sizes, line spacing, paragraph and list definitions,
% and the use of the \verb|\vspace| command to manually adjust the
% vertical spacing between elements of your work --- is not allowed.

% {\bfseries Your document will be returned to you for revision if
%   modifications are discovered.}

\begin{table*}[h]
    \caption{The results in different study regions and times.}
    \vspace{-0.3cm}
    \begin{tabular}{c|ccc|ccc|ccc|cccc}
        \toprule
        \multirow{2}{*}{\textbf{Model}} & \multicolumn{3}{c|}{FP Observed}&\multicolumn{3}{c|}{GoM-10} &\multicolumn{3}{c|}{GoM-11}&\multicolumn{3}{c}{GoM-12} \\  \cmidrule(r){2-4}  \cmidrule(r){5-7}  \cmidrule(r){8-10} \cmidrule(r){11-13} 
        & \textbf{MAE} &\textbf{RMSE} &\textbf{MAPE}& \textbf{MAE} &\textbf{RMSE} &\textbf{MAPE}& \textbf{MAE} &\textbf{RMSE} &\textbf{MAPE}& \textbf{MAE} &\textbf{RMSE} &\textbf{MAPE} \\ \midrule
        \textbf{Kriging} & 16.3373 & 20.8742 & 43.27\% & 0.8235 & 1.9499& 2.50\%& 1.0416& 1.7759& 3.26\%& 1.5211& 1.8672& 2.50\%\\
        \textbf{GWR} & 8.8916 & 19.1230 & 34.16\%  & 2.2176 & 6.8032 & 6.90\% & 1.1731 & 2.9864 & 3.50\% & 0.8039 & 1.2283 & 2.30\%\\
        \hline
        \textbf{MLP} & 2.4257 & 3.7254 & 8.38\%  & 0.3871 & 0.5031 & \textbf{1.15\%} & 0.3525 & 0.5436 & \textbf{1.03\%} & \textbf{0.3696} & 0.7539 & \textbf{1.07\%} \\ 
        \textbf{LSTM} & 3.0902 & 4.7452 & 10.57\% & 0.3980 & 0.5848 & 1.20\% & 0.3814 & 0.6539 & 1.13\% & 0.4679 & 0.8769 & 1.36\% \\
        \textbf{GAN}  &2.8360  &4.4835  &10.26\% & 0.4977& 0.5742  & 1.63\% &  0.8586 & 1.0717 & 2.49\% & 0.4707 &  0.7883 & 1.36\%\\
        \hline
        \textbf{\method}~(\textbf{Ours}) & \textbf{2.3331}  &\textbf{2.9338}  &\textbf{6.83\%} &\textbf{0.3288} &\textbf{0.4247} &  1.29\% &\textbf{0.3088} &  \textbf{0.4084} & 1.21\% & 0.4761 & \textbf{0.6395} & 1.37\%\\
        \bottomrule       
    \end{tabular}
    \vspace{-0.3cm}
    \label{tab:performance}
\end{table*}
\section{Experiments}\label{sec:experiments}

\subsection{Datasets}
We evaluate \method\ on two drifter datasets filtered to retain only coastal and nearshore data. Their key statistics are summarized in Table~\ref{tab:dataset}, with details as follows:

\noindent \textbf{FP Observed} is a real-world drifter dataset collected by the Harbor Branch Oceanographic Institute (HBOI) at Florida Atlantic University. Data were over six non-consecutive days using 2–3 drifters per day. Deployment and retrieval times varied across days, yielding a total of 109 trajectories. After preprocessing to remove anomalies, we retain three valid days (Dec 8, 2015; Dec 15, 2015; and Jun 16, 2016), resulting in 36 trajectories with 17,475 data points. For this study, we focus on Jun 16, 2016, which includes 4 trajectories with 11,426 salinity measurements near Fort Pierce Inlet, Florida, USA.

\noindent{\textbf{GoM Simulated}} is a synthetic drifter dataset generated by HBOI using a numerical ocean current model. The full dataset covers the Gulf of Mexico (Gulf of America) and adjacent waters, comprising 200 simulated drifter trajectories with 2,569 data points each, spanning from Oct 1, 2021, to Jan 16, 2022. To support our focus on salinity interpolation in sparsely sampled regions, we extract a subset near the Mississippi River delta, containing 69 trajectories and 52,253 time steps. This subset is divided into three monthly segments, \textbf{GoM-10} (Oct), \textbf{GoM-11} (Nov), and \textbf{GoM-12} (Dec).

\subsection{Baselines}
We compare~\method~ with five representative baselines spanning statistical and deep learning methods to evaluate its performance:

\noindent \textbf{Kriging}~\cite{chen2016optimization}: A classical geostatistical interpolation method that estimates unobserved values based on spatial autocorrelation. It is widely applied in environmental modeling.

\noindent \textbf{Geographically Weighted Regression (GWR)}~\cite{brunsdon1996geographically}: A spatial regression model that fits local linear models at each location, capturing spatially varying relationships.

\noindent \textbf{Multilayer Perceptron (MLP)}: A feedforward neural network that models nonlinear mappings from spatiotemporal features to salinity values.

\noindent \textbf{Long Short-Term Memory (LSTM)}~\cite{hochreiter1997long}: A recurrent neural network designed to capture temporal dependencies, improving the model's ability to reflect salinity dynamics over time. 

\noindent \textbf{Generative Adversarial Network (GAN)}~\cite{goodfellow2014generative}: A generative model with a discriminator and generator trained adversarially. The generator learns to produce salinity fields conditioned on spatiotemporal inputs, capturing complex high-order patterns.

%These baselines offer a diverse set of interpolation and prediction strategies, enabling a comprehensive evaluation of~\method~in terms of spatial continuity, temporal dynamics, and nonlinear modeling capacity.

\subsection{Evaluation Settings}
\textbf{Dataset.}
%To capture periodic variations in salinity, we encode temporal features using sine and cosine transformations of hour, minute, day, and month:
%\begin{equation}
%\text{hour\_sin} = \sin(2\pi \cdot \text{hour} / 24), \quad \text{hour\_cos} = \cos(2\pi \cdot \text{hour} / 24),  \text{etc.}
%\end{equation}
%This encoding allows the model to learn cyclical patterns.
For each dataset, trajectories are randomly split into training (70\%), validation (15\%), and testing (15\%) sets. To ensure consistent comparisons across all models, the random seed is fixed to 42, so that the same trajectories are used in each split.

%\subsubsection{Implementation Details}
%Kriging and GWR are implemented using the PyKrige and mgwr libraries, respectively. Since GWR cannot handle identical spatial %locations due to numerical instability, we add small random jitters to duplicate coordinates. This preserves spatial structure %while ensuring stable local regression.
%All deep learning models are implemented in PyTorch. The GAN model is trained for up to 300 epochs using the Adam optimizer with a learning rate of $1 \times 10^{-4}$. We employ early stopping with a patience of 10 epochs based on validation loss to prevent overfitting. 
\textbf{Metrics.}
Model performance is evaluated using three regression metrics: 
Mean Absolute Error (MAE), $\frac{1}{N} \sum_{i=1}^{N} |y_i - \hat{y}_i|$; 
Root Mean Squared Error (RMSE), $\sqrt{\frac{1}{N} \sum_{i=1}^{N} (y_i - \hat{y}_i)^2}$; and 
Mean Absolute Percentage Error (MAPE), $\frac{100\%}{N} \sum_{i=1}^{N} \left| \frac{y_i - \hat{y}_i}{y_i} \right|$.
% These metrics are defined as:
% \begin{equation}
% \text{MAE} = \frac{1}{N} \sum_{i=1}^{N} |y_i - \hat{y}_i|
% \end{equation}
% \begin{equation}
% \text{RMSE} = \sqrt{ \frac{1}{N} \sum_{i=1}^{N} (y_i - \hat{y}_i)^2 }
% \end{equation}
% \begin{equation}
% \text{MAPE} = \frac{100\%}{N} \sum_{i=1}^{N} \left| \frac{y_i - \hat{y}_i}{y_i} \right|
% \end{equation}

\begin{figure}[h]
    \centering
    \vspace{-0.2cm}
    \includegraphics[width=0.5\textwidth]{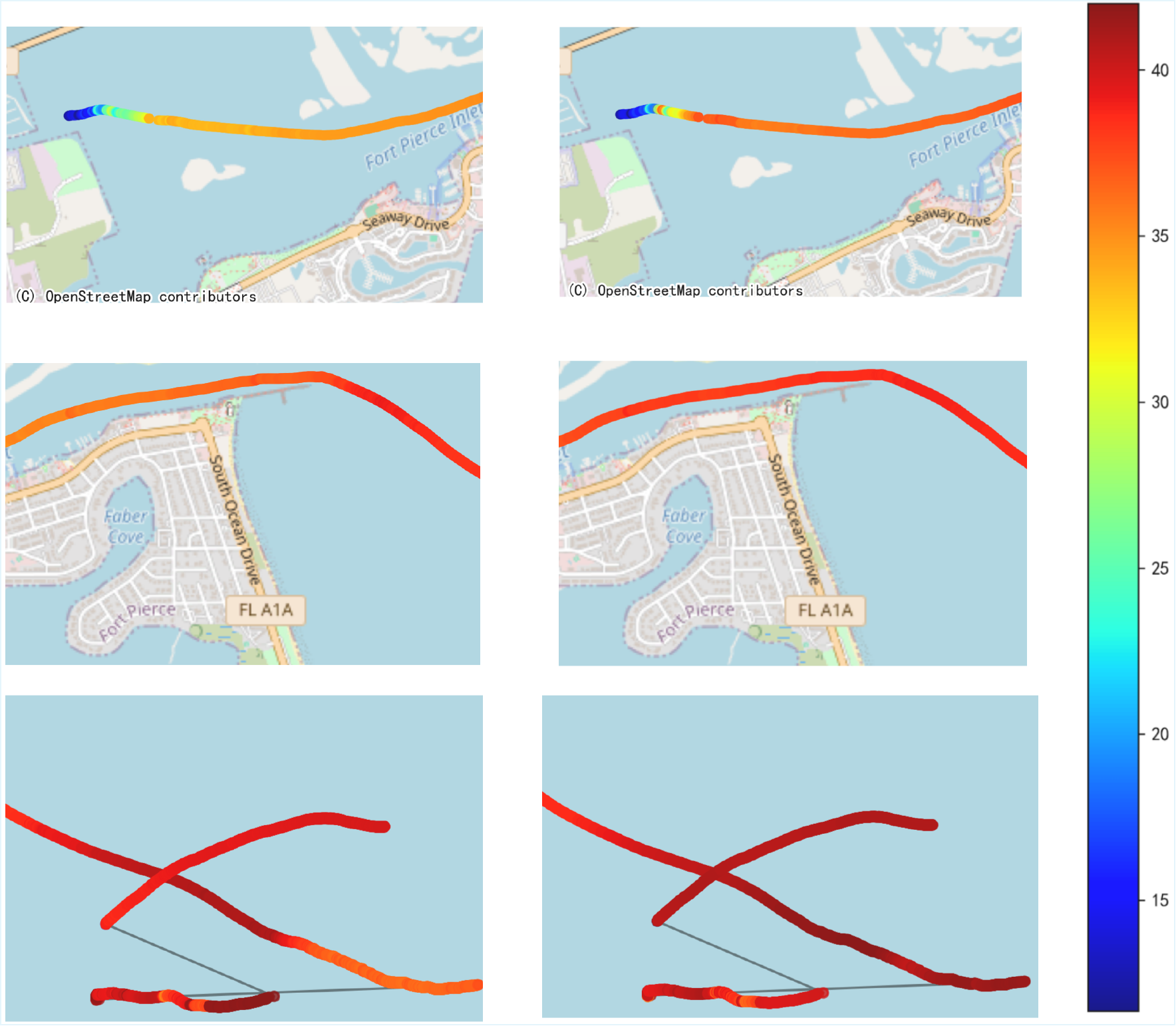}
    \vspace{-0.2cm}
    \caption{Comparison of sea surface salinity (psu) along drifter tracks in the FP Observed area.  \textbf{Left:} True salinity measurements and  \textbf{Right:} \method‐imputed salinity fields. }
    \vspace{-0.2cm}
    \label{fig:Fb-observed}
\end{figure}
\subsection{Quantitative and Visual Results}
Table~\ref{tab:performance} presents a comprehensive comparison of key performance metrics between \method~and various baseline methods. \method~consistently achieves superior performance, especially on the real-world FP Observed dataset. Compared to the best-performing baseline (MLP), \method~reduces MAE by 3.8\%, RMSE by 21.3\%, and MAPE by 18.5\%, highlighting its robustness and imputation accuracy.

Traditional methods such as Kriging exhibit the highest errors, highlighting their limitations in handling complex ocean dynamics. Although learning-based baselines exhibit improved performance, they still fall short in capturing intricate spatiotemporal dependencies, particularly under real-world data sparsity and noise. In some GoM datasets, MLP performs comparably well in MAPE, likely due to the smoother nature of synthetic trajectories, which can favor simpler architectures. However, these isolated gains do not generalize to real settings. In contrast, ~\method~consistently delivers strong results across both real and synthetic datasets, demonstrating its ability to model intricate spatiotemporal structures. 

Figure~\ref{fig:Fb-observed} visualizes salinity distributions in the FP Observed dataset, comparing (a) ground truth and (b)~\method’s imputation. Both plots share the same projection and color scale. The reconstructed field by~\method~accurately reflects the increasing salinity gradient from inlet (blue–green) to nearshore (orange–red), validating its ability to capture large-scale spatial patterns despite sparse sampling. Overall, the results confirm that ~\method~is a robust, generalizable solution for nearshore salinity imputation.

\begin{figure}[h]
    \centering
    \includegraphics[width=0.95\linewidth]{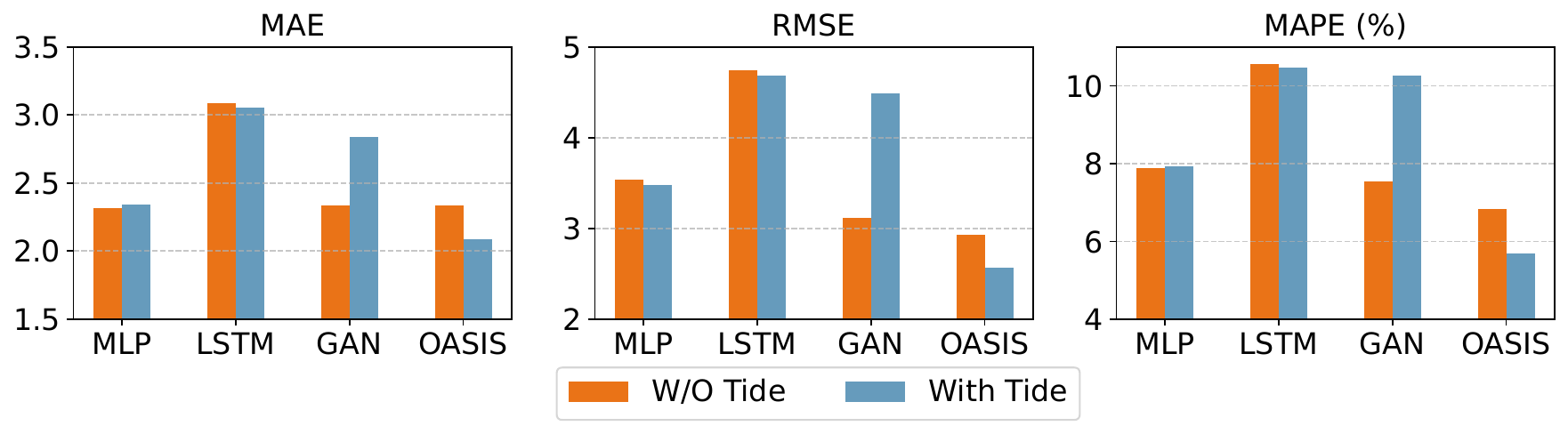}
    \caption{Performance comparison of different models on the FP Observed dataset with and without the tide feature.}
    \vspace{-0.3cm}
    \label{fig:comparison}
\end{figure}

\subsection{Covariate Analysis}
To estimate salinity more accurately, certain physical covariates, such as velocity, sea surface temperature, wind speed, and density, can be leveraged via control equations~\cite{mcdougall2021interpretation}. However, these variables typically require in situ measurements and are thus impractical in many nearshore deployments. To address this limitation, we incorporate tidal height as a low-cost, easily accessible proxy, retrievable from NOAA databases or astronomical models without additional instrumentation. For each dataset, we locate the nearest NOAA Tides and Currents station and retrieve the corresponding tide information. Typically, 2–4 tide events are recorded per day. A sinusoidal function is fitted, either daily or monthly, based on these events, enabling continuous estimation of tide levels for each timestamp in the dataset. 
Figure~\ref{fig:comparison} reports the imputation performance on the FP Observed dataset with and without the tide feature. We observe three main findings:
\textbf{(1) Marginal gains for simple regressors.} Adding tide to MLP yields only minor RMSE reductions, with negligible MAE or MAPE improvement. Similar trends are seen for LSTM, indicating insufficient capacity to exploit the periodic tidal signal. \textbf{(2) Inconsistency in vanilla GAN.} Naïvely appending tide to the standard GAN degrades performance across all metrics. This suggests that without an explicit mechanism to integrate the covariate, the added feature acts as noise, disrupting the adversarial learning dynamics. \textbf{(3) Effective integration in our Diffusion Adversarial Network variant.} By contrast, our framework leverages tide information through a dedicated conditioning module, yielding substantial improvements. These gains demonstrate that tide, as an easily observed proxy for unmeasured covariates, can be effectively exploited by models designed to capture its periodic structure.

In summary, while tide feature provide minimal benefit to generic models and may mislead unstructured GANs, our design successfully integrates this covariate, demonstrating both improved accuracy and robustness in salinity imputation.
%\begin{table}[hbt!]
%\caption{The result in FP Observed dataset with tide feature.}
%\label{tab:tide}
%\begin{tabular}{l|c|c|c}
%\toprule
%\textbf{Model}  & \textbf{MAE}  & \textbf{RMSE} & \textbf{MAPE}    \\ 
%\midrule
%{MLP}              & 2.4257 & 3.7254 & 8.38\%  \\ 
%{MLP + Tide}       & 2.3415 & 3.4807  & 7.93\%  \\
%{LSTM}             & 3.0902 & 4.7452 & 10.57\% \\ 
%{LSTM + Tide}       & 3.0571 & 4.6858 & 10.48\% \\ 
%{GAN}              &2.3343  &3.1120  &7.55\%   \\
%{GAN + Tide}         &2.8360  &4.4835  &10.26\%  \\
%\hline
%\textbf{\method}~         &2.3331  &2.9338  &6.83\%   \\ 
%\textbf{\method}~ \textbf{+ Tide}    & \textbf{2.0864} &\textbf{2.5675}  &\textbf{5.68\%}   \\
%\bottomrule
%\end{tabular}
%\end{table}
\begin{table}[h]
\caption{The result of ablation on FB Observed dataset. Norm, GDC, and SD are normalization, Global Dependency Capturing, and Scheduler Diffusion model, respectively.}
    \centering
    \vspace{-0.2cm}
    \begin{tabular}{c|c|c|c}
    \toprule
         \textbf{Method}& \textbf{MAE}& \textbf{RMSE}& \textbf{MAPE}  \\
         \midrule
         \textbf{\method}&\textbf{2.3331}&\textbf{2.9338}&\textbf{6.83\%}\\
         \textit{w/o} Norm& 2.4522& 3.2500& 7.15\%\\
         \textit{w/o} GDC& 2.7115&3.6371 &8.88\% \\
         \textit{w/o} SD&2.5280 &3.4546 &8.46\% \\
         \bottomrule
    \end{tabular}
    \vspace{-0.2cm}
    \label{tab:ablation}
\end{table}
\begin{figure*}[t]
    \centering
    \vspace{-0.2cm}
    \includegraphics[width=0.9\linewidth]{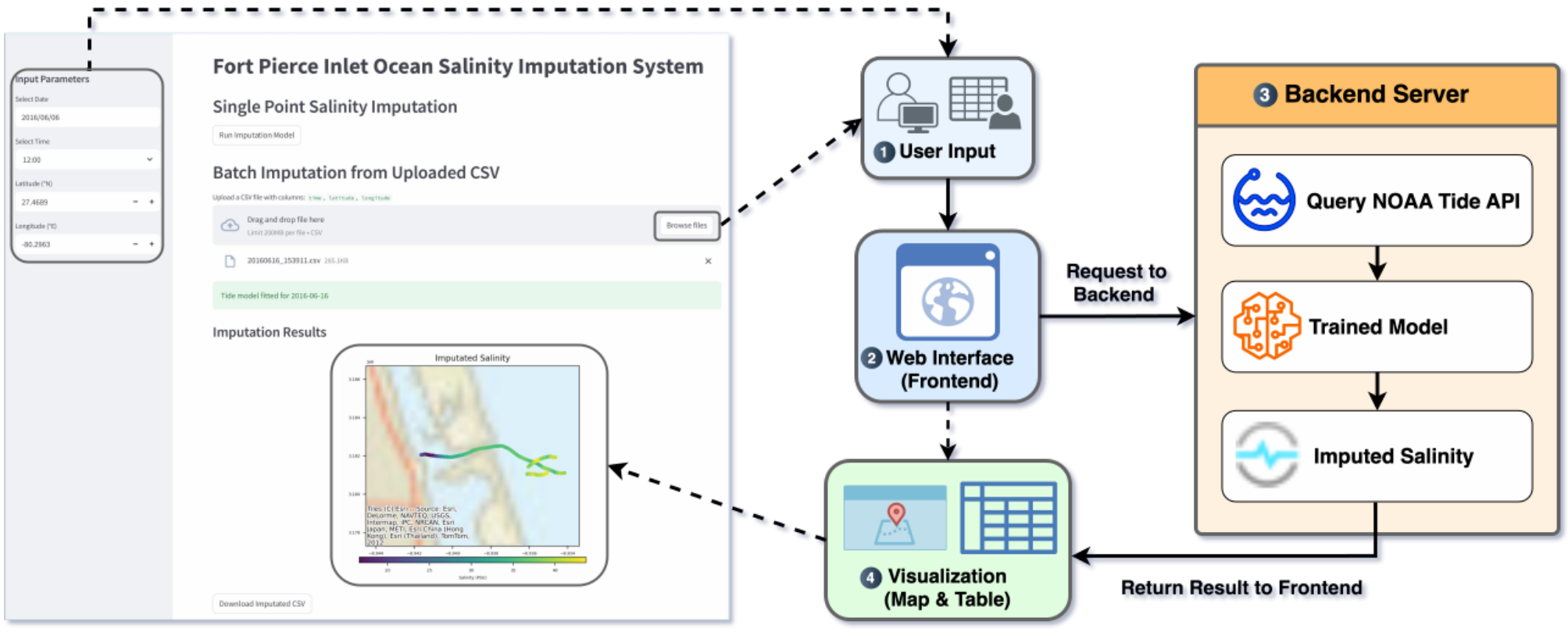}
    \caption{Overview of our~\method: the web-based deployment interface (left) and the underlying system architecture (right).}
    \label{fig:application}
\end{figure*}
\subsection{Ablation Study}
To quantify the contributions of each component in~\method, we conduct an ablation study on the FB Observed dataset (Table~\ref{tab:ablation}). We remove normalization (Norm), the global dependency capturing module (GDC), and the scheduler diffusion model (SD) in turn, and report MAE, MSE, and MAPE. \textbf{(1) Without Norm.}  Omitting the normalization step increases MAE from 2.3331 to 2.4522 (+5.1\%), RMSE from 2.9338 to 3.2500 (+10.8\%), and MAPE from 6.83\% to 7.15\%. This highlights the importance of feature scaling in stabilizing training and reducing bias. \textbf{(2) Without GDC.}  Removing the global dependency capturing module leads to the largest performance degradation (MAE = 2.7115, +16.3\%; RMSE = 3.6371, +24.0\%; MAPE = 8.88\%), demonstrating that modeling long-range spatial–temporal correlations is critical for accurate salinity imputation. \textbf{(3) Without SD.}  Excluding the scheduler diffusion component also worsens results (MAE = 2.5280, +8.4\%; RMSE = 3.4546, +17.7\%; MAPE = 8.46\%), indicating that progressive refinement via diffusion scheduling effectively enhances the generator’s output quality.

These results confirm that each module—normalization, global dependency capturing, and scheduler diffusion contributes substantially to the overall performance of~\method. 
% \subsection{Visualization}
% Figure~\ref{fig:Fb-observed} presents the spatial salinity distribution along drifter tracks in the FP Observed area, comparing (a) the true measurements and (b) our~\method~imputations on the same map projection and color scale. Both plots exhibit a clear increasing salinity gradient from the inlet (blue–green) toward the open ocean (orange–red). Our imputed field closely follows the true gradient, indicating that~\method~accurately captures large‐scale spatial trends.

% The ``\verb|acmart|'' document class requires the use of the
% ``Libertine'' typeface family. Your \TeX\ installation should include
% this set of packages. Please do not substitute other typefaces. The
% ``\verb|lmodern|'' and ``\verb|ltimes|'' packages should not be used,
% as they will override the built-in typeface families.

% The title of your work should use capital letters appropriately -
% \url{https://capitalizemytitle.com/} has useful rules for
% capitalization. Use the {\verb|title|} command to define the title of
% your work. If your work has a subtitle, define it with the
% {\verb|subtitle|} command.  Do not insert line breaks in your title.

% If your title is lengthy, you must define a short version to be used
% in the page headers, to prevent overlapping text. The \verb|title|
% command has a ``short title'' parameter:
% \begin{verbatim}
%   \title[short title]{full title}
% \end{verbatim}

\section{Application Deployment}
To demonstrate the practical utility of \method, we developed a web-based deployment system for on-demand salinity imputation near Fort Pierce Inlet, as shown in Figure~\ref{fig:application}. The interface supports both single-point and batch imputation via interactive input of specific timestamps and coordinates or CSV upload.

The system is implemented with Streamlit (frontend) and PyTorch (backend). Upon receiving user input, the system queries the NOAA CO-OPS API to retrieve tide predictions for Fort Pierce Inlet (station ID: 8722212) on the specified date. A sinusoidal model is then fitted to the daily tide data and used to estimate the tide level at each queried timestamp. 

The estimated tide levels, along with the spatiotemporal metadata, are passed into the pre-trained model to generate imputed salinity values at each target location. The results are presented through a table and a geospatial visualization, where salinity levels are represented using a continuous color scale. The backend is modular and version-controlled. Updated models can be integrated into the deployment pipeline simply by replacing the serialized PyTorch model file (.pt) and its corresponding scaler (.pkl) without modifying the application logic. This enables continuous refinement of the imputation capability without disrupting the user experience.

The current deployment is lightweight and suitable for prototyping and small-scale use. To support broader accessibility and scalability, future versions will migrate to a more robust web architecture (e.g., React or Flask-React) with asynchronous interaction capabilities. The system will also be deployed in cloud environments (e.g., AWS) using Docker to enable scalable, containerized services.

\section{Conclusion}
In this paper, we presented ~\method, a novel diffusion adversarial framework for sea salinity imputation that unifies reversible instance normalization, transformer‐based global dependency modeling, and a scheduler‐guided Adversarial Diffusion Network. By leveraging easily observed tidal height as an auxiliary covariate, ~\method~ is able to capture both large‐scale gradients and fine‐scale temporal fluctuations without requiring specialized instrumentation. Extensive quantitative experiments show consistent improvements over classical and neural baselines across benchmarks. Ablation studies confirm the essential roles of normalization, global dependency capturing, and the diffusion scheduler. Qualitative visualizations further illustrate ~\method’s ability to reconstruct smooth, continuous salinity fields and preserve localized extrema under severe data sparsity. To support real-world use, a lightweight web-based system is developed to enable real-time salinity imputation via interactive and batch inputs, with integrated NOAA tide access and future support for cloud scalability.

\section*{Acknowledgments and Disclosure of Funding}
%    This research is supported by the CSIRO–National Science Foundation (US) AI Research Collaboration Program: Towards Interpretable and Responsible Graph Modeling for Dynamic Systems (IIS-2302786). This work has also been supported by the Australian Research Council (ARC) under grants FT210100097 and DP240101547.

    This research is supported by the U.S. National Science Foundation and Australia CSIRO joint project: NSF-CSIRO: Towards Interpretable and Responsible Graph Modeling for Dynamic Systems (IIS-2302786). This work has also been supported by the Australian Research Council (ARC) under grants FT210100097 and DP240101547.

\section*{GenAI Usage Disclosure}
The authors confirm that all core research activities—including study design, data collection, analysis, and interpretation—were conducted without the use of generative artificial intelligence tools. Generative AI was employed only in a limited capacity during manuscript preparation: specifically for debugging minor coding errors, and for grammar, spelling, and stylistic checks. No substantive content, scientific ideas, or interpretations were generated by AI, and all results, figures, and methodological descriptions reflect the authors’ own work and expertise.

\bibliographystyle{ACM-Reference-Format}
\bibliography{NSF-CSIRO}

\appendix









\end{document}